# The Temporal Logic of Causal Structures

**Samantha Kleinberg**
Courant Institute of Mathematical Sciences
New York University, NY
samantha@cs.nyu.edu

**Bud Mishra**
Courant Institute of Mathematical Sciences
New York University, NY
mishra@nyu.edu

## Abstract

Computational analysis of time-course data with an underlying causal structure is needed in a variety of domains, including neural spike trains, stock price movements, and gene expression levels. However, it can be challenging to determine from just the numerical time course data alone what is coordinating the visible processes, to separate the underlying prima facie causes into genuine and spurious causes and to do so with a feasible computational complexity. For this purpose, we have been developing a novel algorithm based on a framework that combines notions of causality in philosophy with algorithmic approaches built on model checking and statistical techniques for multiple hypotheses testing. The causal relationships are described in terms of temporal logic formulæ, reframing the inference problem in terms of model checking. The logic used, PCTL, allows description of both the time between cause and effect and the probability of this relationship being observed. We show that equipped with these causal formulæ with their associated probabilities we may compute the average impact a cause makes to its effect and then discover statistically significant causes through the concepts of multiple hypothesis testing (treating each causal relationship as a hypothesis), and false discovery control. By exploring a well-chosen family of potentially all significant hypotheses with reasonably minimal description length, it is possible to tame the algorithm's computational complexity while exploring the nearly complete search-space of all prima facie causes. We have tested these ideas in a number of domains and illustrate them here with two examples.

## 1   INTRODUCTION

Work on time series data has generally focused on identifying groups of similar or co-regulated elements using clustering techniques (Bar-Joseph, 2004) or finding patterns via data mining (Agrawal and Srikant, 1995) but comparatively little has been done to infer causal relationships between the elements of these time series. When trying to decipher the underlying structure of a system, what we would ultimately like to know are the rules governing its behavior. That is, knowing not simply its patterns of activity, but what is responsible for this activity would lead to a richer understanding of the system as well as the ability to better predict future events.

In biological systems, one example of this is discovering dependencies between genes, finding those that influence others. These networks provide a model of biological processes that can then be tested and validated using knock-out experiments. Work in this area has primarily used graph based methods (Friedman et al., 2000; Spirtes et al., 2001; Murphy and Mian, 1999), such as Bayesian networks, which provide a less expressive framework than temporal logic in terms of the relationships they can represent and discover. We seek to enrich the existing frameworks in order to enable one to infer, describe and analyze arbitrarily complex causal relationships, which must involve temporal operators as well as probabilistic relationships and propositional connectives. We present a method for this purpose using a probabilistic temporal logic, which allows us to express a richer variety of causal relationships in a compact way. By posing our questions about how two events are related in terms of model checking and inference, we are able to efficiently infer relationships in large datasets with many variables. We will first introduce the problem of causality in philosophy and then discuss graphical methods for causal inference before describing our method and illustrating its use with two different data sets.

## 2  BACKGROUND

The study of causality has a long history in philosophy, going back as far as Ancient Greece. Since then there have been a number of accounts, with the primary ones being in terms of regularities, processes, counterfactuals and as probabilistic relations. Most modern accounts of causality may be traced back to Hume's regularities of observation, where $C$ causes $E$ if whenever $C$ happens, $E$ follows, and we detect this relationship through observation of $C$ and $E$. However, Hume also restated this as "if $C$ had not happened, $E$ would not have happened." That statement is what is known as a counterfactual. These previous theories do not necessarily restrict the cause and effect to be contiguous or even nearby in space. Process theories restate the problem in terms of transfer of conserved quantities (such as momentum), where this transfer requires locality in space-time. However, when it comes to inference, we need some notion of probability, since few relationships are actually deterministic.

The central idea of probabilistic causality is that a cause is something that is earlier than its effect (temporal priority) and that raises the probability of this effect (probability raising). One primary theory comes from Suppes (1970). In Suppes theory, an event $C_{t'}$ is a *prima facie* cause of an event $E_t$ when all of the following conditions hold:

1. $t' < t$,
2. $P(C_{t'}) > 0$, and
3. $P(E_t|C_{t'}) > P(E_t)$.

Note that these formulæ simply encode temporal priority and probability raising, with the stipulation that the cause has a nonzero probability.

However, this simple definition leads to many situations being erroneously labeled causal. If two events ($d$ and $e$) have a common cause ($c$), with $d$ being caused slightly earlier than $e$, we may be led to conclude that $d$ causes $e$. This situation arises because whenever $c$ happens, it raises the probability of both $d$ and $e$, but since $d$ is slightly earlier than $e$, it too appears to raise the probability of $e$. It is not enough, then, to look only for something that raises the probability of the effect, as this erroneously admits relationships that are not truly causal. To address this problem, there have been a number of approaches, which may be roughly divided into two groups: those that look for common causes or earlier events that account at least as well for the effect (Suppes, 1970; Reichenbach, 1956); and those that look at sets of background factors, or contexts, and test whether the cause raises the probability of the effect with respect to these contexts (Eells, 1991). In the first case, approaches such as that of Suppes look for a third event ($D_{t''}$) that occurs earlier than the prima facie cause ($C_{t'}$) and the effect ($E_t$), such that:

1. $P(C_{t'} \wedge D_{t''}) > 0$,
2. $|P(E_t|C_{t'} \wedge D_{t''}) - P(E_t|D_{t''})| < \epsilon$, and
3. $P(E_t|C_{t'} \wedge D_{t''}) \geq P(E_t|C_{t'})$.

That is, $C_{t'}$ and $D_{t''}$ must occur together with nonzero probability, and $C_{t'}$ makes less than some small $\epsilon$ difference to the probability of $E_t$, once $D_{t''}$ is known. If all of these conditions are true, then $C_{t'}$ is called an $\epsilon$-*spurious cause* of $E_t$. Note that an event at the same time as $C_{t'}$ or at a time between $C_{t'}$ and $E_t$ cannot make $C_{t'}$ spurious under this method. There may be causes - themselves not caused by $C_{t'}$ - that occur temporally between $C_{t'}$ and $E_t$ that will not be considered. However, the primary problem is that of finding an appropriate value for $\epsilon$. Further, even after finding the "correct" value, there need only be one such $D_{t''}$ for $C_{t'}$ to be deemed $\epsilon$-spurious, while there may be hundreds of other factors for which it makes a significant positive difference.

Another approach, that of Eells (1991), is to compute the average significance of a cause for its effect. In other words, one does not just seek any single more powerful cause, but rather measures, overall, how well the cause predicts the effect. First, Eells defines the set of causal background contexts, $\{K_1, K_2...K_n\}$. These are formed by holding fixed the set of all factors causally relevant to the effect (that occur at any time earlier than the effect) in all possible ways. For instance, given a set of three factors, $\{x_1, x_2, x_3\}$, one possible background context would be $\{x_1, \neg x_2, x_3\}$. Eells then defines the average degree of causal significance (ADCS) of a factor $C$ for a factor $E$:

$$\sum_i Pr(K_i)[Pr(E|K_i \wedge C) - Pr(E|K_i \wedge \neg C)]. \quad (1)$$

By taking the average difference $C$ makes to $E$'s probability in each context, weighted by the probability of that background context occurring, we can weed out artifacts that make a consistent but small difference to the probability of the effect.

In contrast to Suppes, Eells considers events that occur at any time prior to the effect (versus only those prior to the potential cause). Also, instead of looking for single causes that account better for the effect, Eells considers contexts. The result is not a partition into genuine/spurious causes, but rather a quantity denoting how significant each cause is for its effect. As before, it is unclear how to treat the number produced by the ADCS and determine which values are significant.

## 3 PROBLEM OVERVIEW

While the philosophical theories described give us an idea of how to identify causes, it still remains to turn this into an automated algorithmic approach. The first work in automating the inference of causal relationships was by Pearl (2000) and Spirtes et al. (2000) (SGS), using graphical models, such as Bayesian networks (BNs). In these approaches, the causal structure of the system is represented as a graph, where variables are represented by nodes and the edges between them represent conditional dependence (and the absence of an edge implies conditional independence). Then, a number of assumptions about the data can be used to direct these edges from cause to effect. The result is a directed acyclic graph (DAG) where a directed edge between two nodes means the first causes the second. In these graphical approaches, the edges can be oriented without the use of time course data, as a consequence of the other assumptions. We will use the terminology of SGS and primarily describe their work, though these assumptions and the general procedure are used by many with some variation.

First, the *Causal Markov condition* (CMC) is that a node in the graph (variable) is independent of all of its non-causal descendants given its direct causes (those that are connected to the node by one edge). The inference of causal structures relies on two more assumptions: faithfulness and causal sufficiency. *Faithfulness* assumes that *exactly* the independence relations found in the causal graph hold in the probability distribution over the set of variables. This requirement implies that the independence relations obtained from the causal graph are due to the causal structure generating it. If there are independence relations that are not a result of CMC, then the population is unfaithful. The idea of faithfulness is introduced to ensure that independencies are not from chance coincidence but from some genuine structure. *Causal sufficiency* assumes that the set of measured variables includes all of the common causes of pairs on that set. In cases where causal sufficiency does not hold, then the inferred graphs can include those with nodes representing unmeasured common causes that could also have lead to the observed distribution. Knowledge about temporal ordering can also be used at this point if it is available. In general the conditional independence conditions are assumed to be exact conditional independence, though it is possible to define some threshold to decide when two variables will be considered independent. The result is a set of graphs that all represent the independencies in the data, where the set may contain only one graph in some cases when all assumptions are fulfilled. However, when using these graphical models there is no natural way of representing or inferring the time between the cause and the effect or a more complex relationship than just one node causing another at some future time.

Following the use of BNs, dynamic Bayesian networks (DBNs) (Friedman et al., 1998) were introduced to address the temporal component of these relationships. DBNs extend BNs to show how the system evolves over time. For this purpose, they generally begin with a prior distribution (described by a DAG structure) as well as two more DAGs: one representing the system at time $t$ and another at $t+1$, where these hold for any values of $t$. The connections between these two time slices then describe the change over time. As before, there is usually one node per variable, with edges representing conditional independence. Note that this implies that while the system can start in any state, after that the structure and dependencies repeat themselves. That is, the relationships from time 10 to 11 are exactly the same as those from time 11 to 12.

Another statistical method, applied primarily to economics, was developed by Granger (1969) to take two time series and determine whether one predicts, or causes, the other with some lag time between them. Building on this, recent work by Eichler and Didelez (2007) focuses on time series and explicitly capturing the time elapsed between cause and effect. They define that one time series causes another if an intervention on the first alters the second at some later time. That is, there may be lags of arbitrary length between the series, and they find these lags as part of the inference process. While it is possible to also define the variables in this framework such that they represent a complex causal relationship as well as the timing of the relationship, the resulting framework still does not easily lead to a general method for testing these relationships. Further, while DBNs are a compact representation in the case of sparse structures, it can be difficult to extend them to the case of highly dependent data sets with thousands of variables, none of which can be eliminated.

When we talk about one thing causing another, particularly in terms of scientific data, rarely is it as simple as "$a$ causes $b$", deterministically, with no other relevant factors. Recent work by Langmead et al. (2006) has described the use of temporal logic for querying pre-existing DBNs, by translating them into structures that allow for model checking. This approach allows the use of known DBNs for inference of relationships described by temporal logic formulæ. However, only a subset of DBNs may be translated in this

way(Langmead, 2008), and thus the benefit of this approach (as opposed to one where the model inferred already allows for model checking) is limited.

In contrast to the approaches described earlier, we seek a unified framework that captures, at equal levels, both the probabilistic dependencies and temporal priorities inherent in causal relationships. We motivate our approach with an example considering the temporal and probabilistic relationship between smoking and lung cancer. Cigarettes in the UK currently bear the warning "smoking kills," but this statement does not tell us how likely it is that a person who smokes these labeled cigarettes will die from smoking nor how long it will take for death to occur. Given the choice between packages labeled "smoking kills in 90 years" and "smoking kills within 10 years", we might make very different decisions than when confronted with one that simply says "smoking kills." Note that we have not yet mentioned the probability of death in either case. The first case could be with probability 1 with the latter being much smaller. This additional information and the way it may affect our decision making process points to the need for a more detailed description. That is, when we describe a causal relationship, we need to be able to describe its probability and the time over which it takes place.

## 4 METHOD

Our main problem may be formalized as follows: given a set of time series data representing activities of a system in which we hypothesize that there may exist a causal structure, we seek to infer the underlying relationships forming this structure. In order to represent and infer these relationships, which may be far more complex than one event causing another, we use a probabilistic temporal logic. Broadly, the steps of our method will be to represent the possible causal relationships as logical formulæ, use model checking to determine whether they are satisfied by the system, and then determine which of the satisfied relationships are causal using a measure of their degree of significance and false discovery control to determine at what level something is causally significant. Before describing how this representation relates to probabilistic notions of causality, we give a brief overview of temporal logic, and the particular logic used, PCTL.

### 4.1 TEMPORAL LOGIC

Temporal logics modify traditional modal logics to allow description of *when* a formula is true. That is, rather than just "necessity" or "possibility", a formula may be true at the next point in time or at some point in the future. In branching time logics, such as Computation Tree Logic (CTL) (Clarke et al., 1999), the future may be along one of multiple possible paths. As opposed to a linear time temporal logic, for which there is only one possible future path, we can express whether a property holds for *all* possible paths $(A)$, or if there *exists* at least one path for which it is true $(E)$. The truth values of these formulæ are determined relative to a Kripke structure, a graph with a set of states, transitions between states, and labels indicating propositions true within the states.

We will use a probabilistic extension of CTL, Probabilistic Computation Tree Logic (PCTL), as introduced by Hansson and Jonsson (1994), as it allows probabilistic state transitions, as well as explicit deadlines for when a formula must hold. We begin with a set of atomic propositions, $A$, and a structure (called a discrete time Markov chain (DTMC)) $K = \langle S, s_i, L \rangle, \mathcal{T}$, where:

$S$ is a finite set of states;

$s_i$ is an initial state;

$L : S \rightarrow 2^A$ is a state labeling function; and

$\mathcal{T} : S \times S \rightarrow [0, 1]$ is a transition function such that:
$$\forall s \in S \sum_{s' \in S} \mathcal{T}(s, s') = 1.$$

Then, relative to this structure we can define state formulæ (those that hold within a state) and path formulæ (those that hold along some sequence of states):

1. Each atomic proposition is a state formula.

2. If $f_1$ and $f_2$ are state formulæ, so are $\neg f_1$, $f_1 \wedge f_2$, $f_1 \vee f_2$, and $f_1 \rightarrow f_2$.

3. If $f_1$ and $f_2$ are state formulæ, and $t$ is a nonnegative integer or $\infty$, $f_1 U^{\leq t} f_2$ and $f_1 \mathcal{U}^{\leq t} f_2$ are path formulæ.

4. If $f$ is a path formula and $0 \leq p \leq 1$, $[f]_{\geq p}$ and $[f]_{>p}$ are state formulæ.

The "Until" $(U)$ formula in (3) means that the first subformula $(f_1)$ must hold at every state along the path until a state where the second subformula $(f_2)$ holds, which must happen in less than or equal to $t$ time units. The modal operator "Unless" $(\mathcal{U})$ is defined the same way, but with no guarantee that $f_2$ will hold. In that case, $f_1$ must hold for a minimum of $t$ time units. Finally, in (4) we add probabilities to these until and unless path formulæ to make state formulæ. For example, $[f_1 U^{\leq t} f_2]_{\geq p}$ (which may be abbreviated as $f_1 U^{\leq t}_{\geq p} f_2$), means that with probability at least $p$, $f_2$ will become true within $t$ time units and $f_1$ will hold along the path until that happens. The probability of a state formula is calculated over the set of possible paths from the state, where the probability of a path is the product of the transition probabilities along the path, and the probability for a set of paths is the sum of the individual path probabilities. Path quantifiers analogous to those in CTL may be defined by:

$$Af \equiv [f]_{\geq 1},$$
$$Ef \equiv [f]_{>0},$$
$$Gf \equiv f\mathcal{U}^{\leq \infty} false, \text{ and}$$
$$Ff \equiv true\ U^{\leq \infty} f.$$

Finally we will also make use of the "leads-to" operator (Hansson and Jonsson, 1994):

$$f_1 \leadsto_{\geq p}^{\leq t} f_2 \equiv AG[(f_1 \rightarrow F_{\geq p}^{\leq t} f_2)]. \qquad (2)$$

This formula implies that for every path from the current state, if we are in a state where $f_1$ holds then through some series of transitions taking time $\leq t$, with probability $p$, we will finally reach a state where $f_2$ holds. One difficulty with this formulation is that, as defined, "leads-to" also considers the case where $f_1$ and $f_2$ are true at the same state as one satisfying this formula. We will need to stipulate that there must be at least one transition between $f_1$ and $f_2$. In addition to being important for our temporal priority condition for casuality, this is also in keeping with how one naturally reasons about the term "leads to." We thus wish to write:

$$f_1 \leadsto_{\geq p}^{\geq t_1, \leq t_2} f_2, \qquad (3)$$

which is interpreted to mean that $f_2$ must hold in between $t_1$ and $t_2$ time units with probability $p$. If $t_1 = t_2$, this case simply says it takes exactly $t_1$ time units for $f_2$ to hold. Proofs that this lower bound may be added will appear in an extended version of this paper.

Verifying whether these formulæ hold for a particular system is the goal of model checking, where a system is said to satisfy a formula if its initial state satisfies the formula. For more information on this process, we refer the reader to the original paper by Hansson and Jonsson (1994) as well as a later paper by Baier et al. (1997) that describes a symbolic approach to the problem.

### 4.2 CAUSES AS LOGICAL FORMULÆ

Our aim is now to describe the conditions for causality and express the causes themselves as PCTL formulæ. Recall that the formulæ are defined with regards to a probabilistic structure, as described earlier. While these graphical structures consist of labeled nodes and edges, note that unlike the causal models previously described, such as BNs, the arrows between states have no causal interpretation. They only imply that it is possible to transition from the state at the tail to the state at the head with some non-zero probability (which is used to label this edge in diagrams). Note also that there may be multiple states with the same labels. For example, there may be two states labeled with identical sets of propositions, but that are reached by different paths and which have different paths possible from them. Thus, this graphical model fundamentally differs from Bayesian networks, where each variable generally has one node with incoming and outgoing edges (and lack thereof) representing (in)dependencies. Our structures may also contain cycles, allowing for feedback loops.

We will begin by giving the basic conditions for causality, starting with those for *prima facie* causality. We specify the temporal priority condition of the causal relationship in terms of the time that elapses between cause and effect. If $c$ occurs at some time $t'$ and $e$ occurs at a later time $t$, we characterize the relationship by the time that elapses between them, $|t' - t|$. That is, if we want to state that after $c$ becomes true, $e$ will be true with probability at least $p$ in $|t' - t|$ or fewer time units – but with at least one time unit between $c$ and $e$ – we write:

$$c \leadsto_{\geq p}^{\geq 1, \leq |t'-t|} e. \qquad (4)$$

If we simply want $c$ to be earlier than $e$, that upper bound will be infinity. Note that $c$ and $e$ are any valid PCTL formulæ. For example, in the case of biological systems, we could have:

$$(a_{up} \wedge b_{down}) U c_{up} \leadsto_{\geq 0.9}^{\geq 1, \leq 4} d_{up}, \qquad (5)$$

where this formula states that the simultaneous up-regulation of gene $a$ and suppression of gene $b$, until gene $c$ becomes up-regulated, subsequently results in the up-regulation of gene $d$, with probability 0.9, in between 1 and 4 time units.

Consequently, the probabilistic nature of the relationship between cause and effect can be described in terms of the probability of reaching $c$ and $e$ states and of the paths between $c$ and $e$ states. We need to state that it is possible (beginning from the initial state of the system) to reach a state where $c$ is true, and that the probability of reaching a state where $e$ is true (within the time bounds) is greater after being in a state where $c$ is true (probability $\geq p$) than it is by simply starting from the initial state of the system (probability $< p$). Recall that we do not begin with any a priori knowledge of this structure, but rather aim to recreate this from the data. However, our time course observations may be viewed as a sequence of the possible states occupied by the system. From their ordering and frequency, we may find the possible transitions and their probabilities, which characterize a structure as described earlier. We now define *prime facie*, or potential, causes, as shown below.

**Definition 4.1.** $c$ is a prima facie cause of $e$ if the following conditions all hold:

1. $F^{\leq \infty}_{>0} c$,
2. $c \leadsto^{\geq 1, \leq \infty}_{\geq p} e$, and
3. $F^{\leq \infty}_{<p} e$.

Note that this definition inherently implies that there may be any number of transitions between $c$ and $e$, as long as there is at least one, and the sum of the set of path probabilities is at least $p$. We can also further restrict this time window, when background knowledge makes it possible, but the minimum condition is that $c$ is earlier than $e$ by at least one time unit, and raises the probability of $e$.

### 4.3 TESTING FOR SIGNIFICANCE

As we saw earlier, we need a way of finding which of these possible causes are not significant. However, we do not want to deem a cause spurious if there is only one other cause that makes it so, as Suppes does, since there may be multiple genuine causes of an effect with varying strength. With Eells's approach we face the problem that since there are a large number of such contexts 1) it is rare to have enough data to see them occur with high enough frequency to be meaningful, and 2) testing all such contexts is a non-trivial computational task. If each background context occurs with nonzero probability, we will have $2^n$ such contexts, where $n$ is the number of relevant factors. In our examples, where we may have data for thousands of genes, it is not possible to construct such a set of contexts (let alone to do so for each possible cause whose significance we aim to compute). We also have the problem of: Which values of the ADCS (average degree of causal significance) should be considered significant?

We take our inspiration from both of these methods and proceed as follows. To determine whether a particular $c$ as a cause of $e$ is *insignificant*, we compute the average difference in probabilities for each prima facie cause of an effect in relation to all other prima facie causes of the effect. We begin with $X$ being the set of prima facie causes of $e$. Then, for each $x \in X \setminus c$, we compute the predictive value of $c$ in relation to $x$, by comparing the probability of transitioning to an $e$ state from a $c \wedge x$ state versus a $\neg c \wedge x$ state. If these probabilities are very similar, then $c$ might be an insignificant cause of $e$. As noted earlier, there may only be one such $x$, while there may be a number of other $x$'s for which there is a large difference in the computed probabilities. With:

$$\epsilon_x(c, e) = P(e|c \wedge x) - P(e|\neg c \wedge x), \quad (6)$$

we compute:

$$\epsilon_{avg}(c, e) = \frac{\sum_{x \in X \setminus c} \epsilon_x(c, e)}{|X|}. \quad (7)$$

For each prima facie cause, we have now computed its average potency as a predictor of its effect. If there is only one other cause that would make a cause $c$ seem "spurious", but a number of other factors (that are themselves actually spurious causes of the effect), then we will find that $c$ has a high value of this measure. Finally, we use this $\epsilon_{avg}$ to determine $c$'s significance.

**Definition 4.2.** A prima facie cause, $c$, of an effect, $e$, is an $\epsilon$-*insignificant cause* of $e$ if $\epsilon_{avg}(c, e) < \epsilon$.

A prima facie cause that is not $\epsilon$-insignificant is not necessarily genuine. For the moment we will refer to these causes as only *significant*, with the full detail about when such causes may be labeled genuine postponed to a longer future publication.

Here again we come to the issue of what value of $\epsilon$ should be chosen. An appropriate value can be found using knowledge of the problem, through simulation, or based on other statistical tests. Since we are testing a multitude of hypotheses (from thousands to hundreds of thousands), we determine the appropriate value of $\epsilon$ statistically, using methods for false discovery rate (fdr) control. In particular, we apply the empirical Bayesian formulation proposed by Efron (2004). This method uses an empirical rather than theoretical null, which has been proven to be better equipped for cases where the test statistics are dependent —as may be true in the case of complex causal structures.

We have chosen to control the proportion of falsely rejected null hypotheses (when we incorrectly deem a hypothesis significant) rather than the proportion of falsely accepted null hypotheses (when we incorrectly deem a hypothesis insignificant). Since we are testing a large number of hypotheses, we will accept that we may miss some opportunities for discovery, but we want the discoveries we make to be very likely to be legitimate. For example, when we apply these methods to biomedical (say, malaria) data, our ultimate goal is to propose candidate gene or biomarker targets to explore for vaccine development. In that case it is costly to explore each hypothesis, and better to be more certain about those that are proposed.

The basic idea of this approach is that we assume our data contains two classes, namely, significant and in-significant. We assume the significant class is small relative to the insignificant class, and that these correspond to rejection and acceptance of the null hypothesis, with prior probabilities $p_1 = 1 - p_0$ and $p_0$. That is, $p_0$ and $p_1$ are the prior probabilities of a case (here, a causal hypothesis) being in the "insignificant" or "significant" classes respectively, with these probabilities distributed according to an underlying density. For each hypothesis, we have its associated z-value, the number of standard deviations its significance score is from the mean. We define the mixture density as:

$$f(z) = p_0 f_0(z) + p_1 f_1(z), \qquad (8)$$

Then the posterior probability of a case being insignificant given $z$ is:

$$Pr\{null|z\} = p_0 f_0(z)/f(z), \qquad (9)$$

and the *local false discovery rate* (fdr), is:

$$fdr(z) \equiv f_0(z)/f(z). \qquad (10)$$

Note that, in this formulation, the $p_0$ factor is ignored, yielding an upper bound on $fdr(z)$. Assuming that $p_0$ is large (close to 1), this simplification does not lead to massive overestimation of $fdr(z)$. One may also choose to estimate $p_0$ and thus include it in the fdr calculation, making $fdr(z) = Pr\{null|z\}$. Thus, the significance testing portion of the procedure is:

1. Estimate $f(z)$ from the observed $z$-values (for example by a spline fit);
2. Define the null density $f_0(z)$ from the data;
3. Calculate $fdr(z)$ using equation (10).

Overall the steps of our inference process are:

1. Enumerate logical formulæ — using background knowledge or testing those up to a specified level of complexity;
2. Test which of the formulæ are satisfied by the system and satisfy the conditions for prima facie causality;
3. For each prima facie cause, compute the associated $\epsilon_{avg}$;
4. Translate values for $\epsilon_{avg}$ into z-values and compute fdr as above;
5. For each causal hypothesis where $\epsilon_{avg}(c, e)$ corresponds to $z_i$ such that $fdr(z_i)$ is less than a threshold (say, 0.01) label it as significant, and the rest as insignificant.

The advantage of this procedure is that we can test any PCTL formula as a causal hypothesis and do not need to specify an arbitrary threshold on what difference a cause must make to its effect in order to not be considered insignificant. Rather, the meaning of our threshold is concrete and grounded in statistics.

## 5 EXAMPLES

While biological examples, such as that of gene expression data from microarrays, are of great importance, the inference from such cannot be validated against some ground truth i.e., we can never be certain that the inferred relationships correspond truly to the relationships of the system. In order to validate our methods, we started by testing them on synthetic data sets, where the true causal relationships could be revealed after the inference process was completed. Here we discuss one such data set, representing the firings of a set of synthetic neurons over time. We then discuss a data set where the causal structure is not known, but where the story told by the relationships inferred is consistent with what is known about the system.

### 5.1 SYNTHETIC MEA EXPERIMENTS

This data set consists of a series of synthetically generated patterns, and thus we may eventually reveal the assumed true causal neural networks that were embedded in the simulations[1]. The data were created to mimic multi-neuronal electrode array (MEA) experiments, in which neuron firings may be tracked over a period of time. Data was generated for five different structures, with neurons denoted by characters of the English alphabet. Each data set contained 100,000 firings generated with one embedded structure plus a degree of noise (this is a parameter that was varied). In the example shown here, the underlying structure was a binary tree of four levels. At each time point a neuron can fire randomly (dependent on the noise level selected) or may be triggered to fire by one of its cause neurons. Additional background knowledge was known and used by the inference algorithm: there is a 20 unit refractory period after a neuron fires and then a 20 unit window of time after this when it may trigger another to fire. Consequently, our algorithm only needed to search for relationships where one neuron causes another to fire during a window of 20–40 time units after the causal neuron fires. Condition 2 of prima facie causality is then replaced with $c \leadsto_{\geq p}^{\geq 20, \leq 40} e$, where $c$ and $e$ are individual neurons.

The results from all data sets, as well

---

[1] The data was provided as part of the $4^{th}$ KDD workshop on Temporal Data Mining. It is publicly available at: http://people.cs.vt.edu/~ramakris/kddtdm06/.

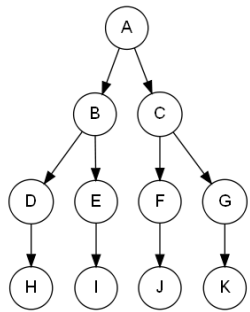

Figure 1: Inferred causal structure, with arrows denoting that the neuron at the tail causes the neuron at the head to fire within 20 to 40 time units with high probability.

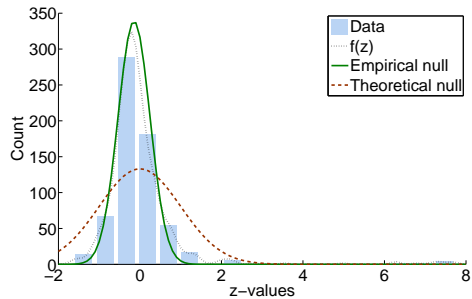

Figure 2: Neural spike train example. We tested pairwise causal relationships, taking into account the known temporal constraints on the system.

as comparison with the PC algorithm and Granger causality are available at http://cs.nyu.edu/~samantha/papers/tlcs.html. Here we examine in detail one of the five structures recovered. Figure 1 shows this, which is one of the most difficult structures to infer, as neurons such as $D$ and $E$ are both highly correlated with $H$ and $I$. The process enumerated 641 prima facie causal hypotheses. We then computed the empirical null from the set of $\epsilon_{avg}$ values using the method and R code made available by Jin and Cai (2006). The results are shown in Figure 2, with a histogram of the computed z-values for the causal hypotheses. The empirical null in this case is given by $N(-0.14, 0.39)$, so it is shifted slightly to the left of the theoretical null, and significantly narrower. The tail of the distribution extends quite far to the right, continuing up to 8 standard deviations away from the mean (almost 20 times the empirical standard deviation). A close up of this area is shown in figure 3. The results obtained here are consistent with the known causal structures that were used to create the simulated data. The ten genuine causal relationships were the only hypotheses with $z$-values greater than three, though there were seven others that, like these, had an fdr of zero.

With no prior knowledge, there are two methods for determining the actual causes. First, in a case where there are few causal relationships found, such as in this example, we can examine the individual hypotheses and manually filter the causal hypotheses. For instance, if there are two causes of an effect, say, one with $z$-value 7.2 and the other with a value 1.3, we may speculate that the former is more likely to be the genuine cause. If the data were experimental, we could do further testing to validate (or refute) this claim. Second, had there been a larger number of prima facie causes of each effect, we could treat each of those as a family of hypotheses, conducting the procedure after the computation of $\epsilon_{avg}$ on each of these families individually. With future research, it may also be possible to better estimate the empirical null distribution.

While we provide the full comparison only on our website and omit it here due to space considerations, we will summarize the results. We used the Tetrad IV implementation of the PC algorithm and the `granger.test` function in the MSBVAR R package, with a lag of 20 time units. We note that our false discovery rate over all tests and false negative rate are: 0.0093 and 0.0005 respectively, while that of the Granger test are: 0.5079 and 0.0026 and PC are: 0.9608 and 0.0159. Further, our discoveries were highly robust, with over 95% of relationships found in both datasets for a particular pattern and noise level.

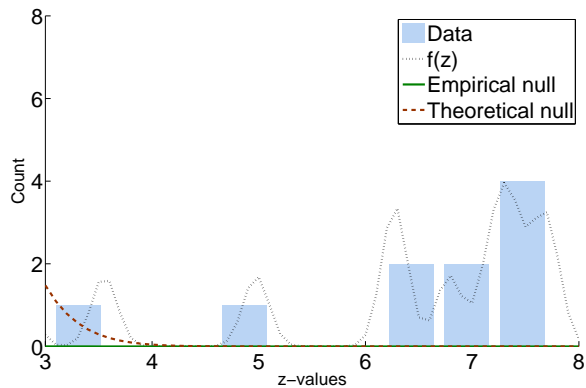

Figure 3: Close-up of the tail area of Figure 2. The relationships in this area are exactly those of Figure 1.

### 5.2 MICROARRAY DATA

To test our method using actual data with an unknown causal structure, we used a data set consisting of a set of time-course gene expression profiles as measured by microarrays. Microarrays allow the mea-surement of expression levels (indicators for whether genes are more or less active than some control) for thousands of genes simultaneously. In a time-course experiment, these levels are measured at regular intervals, giving a picture of how the system behaves over some period of time. The data we examined spans the 48 hour Intraerythrocytic Developmental Cycle of the malaria parasite *Plasmodium falciparum* (Bozdech et al., 2003).

One interesting feature of this segment of the *P. falciparum* life cycle is that all genes are active at some point during its 48-hours. This forms what is referred to as a "cascade" of activity, where as one group of genes become up-regulated, others are down-regulated. We tested relationships between pairs of genes, where the influence occurred at the next time point. That is, we considered formulæ of the form: $c \leadsto_{\geq p}^{\geq 1, \leq 1} e$, where $c$ and $e$ represent the under- or over-expression of particular genes. While we looked at these relationships over the entire timecourse, one could also speculate that different relationships are active at different points during the cycle, and thus require further analysis within each stage of the cycle separately. Since the available timecourse data is comparatively short and sparsely sampled, such finer analysis is likely to result in overfitting, so we decided to focus on the full timecourse data here.

We restricted our testing to only genes whose proteins are involved in known protein-protein interactions, thus leaving out all but 2846 unique genes. To estimate $f(z)$, we used a spline fit to the histogram. The empirical null was estimated using the same method as for the neural spike train data set, and computed to be $N(-1.00, 0.89)$, with the results of our analysis shown in Figure 4. We see that were we to use the theoretical null, this hypothesis would mostly explain our data and we would find very few significant causal relationships. The data itself appears to follow a normal distribution, but with a long right tail. The empirical null is shifted far to the left of the theoretical null, taking into account this tail.

Note that there are thousands of *prima facie* causes with low false discovery rates when using the empirical null hypothesis. This data set, being from an experiment where the structure of the system is not known, does not allow for easy validation of these results. However, we may ask whether they are consistent with what *is* known. In this case, it is commonly believed that biological systems are quite robust, giving rise to backup mechanisms that allow processes to continue uninterrupted in the case of perturbations. We summarize the results from this example as follows: 1) Genes that are active during the same stage of the cycle will have similar patterns of regulation and will be involved in complex sets of relationships occurring in that stage, and 2) Given that we will have a large number of causal relationships orchestrating the cascade, there will likely be a number of back-up mechanisms to allow the uninterrupted cascade of expression. Graphing the set of relationships has shown that there is primarily one network that follows the pattern of the cascade (sets of up-regulatory relationships followed by up-regulation for one set of descendants and down-regulation for another). We could use a very low value (perhaps 0) for the fdr if our goal is a small set of likely relationships (as is the case when our aim is a set of genes to explore for possible vaccine use), as well as speculating two possible classes of causal relationships: genuine and backup causes. Finer analysis of such examples will be dealt with in future work.

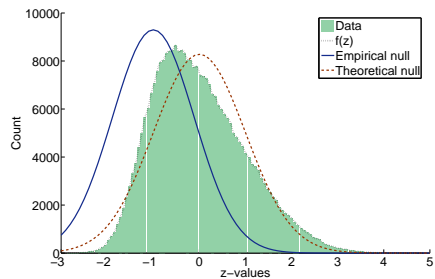

Figure 4: *P. Falciparum* microarray example. We tested causal relationships amongst all pairs of genes over the entire time-course. The histogram shows the number of prima facie causes with a given $z$-value.

## 6 OPEN PROBLEMS

This general method points to a number of open problems. First, we would like to be able to reason about token causality, where the aim is not to find general causal relationships, such as that between smoking and lung cancer, but rather to find the cause on a particular occasion - did a specific person's smoking cause his lung cancer. Secondly, we have assumed as most others do that our time series are stationary, while many systems may have periods of reorganization, where the causal regime switches. Another line of work is to integrate spatial information, perhaps by another modal operator, allowing a notion of not just temporal locality but spatial locality as well. Finally, abstraction has been an important topic in computer science and will have applications in causality. There has not been a natural way of determining at what level one should try to infer causal relationships (e.g. in the case of biology we could look at humans, mechanisms connecting organs, regulation of individual genes, and work our way down to physical relationships), but abstraction may be able to help us answer these questions.

In future work we will address many of these classical problems.

# 7 CONCLUSION

Understanding complex causal structures is at the heart of many disciplines, particularly those of biology and the social sciences, where theories explain interacting parts of a whole and are subject to missing and incomplete data and must be revised as new experiments reveal new structure. There is a rich literature in philosophy on what constitutes a causal relationship and how these may be identified. We have exploited the fact that the conditions given for causality in this philosophical tradition may be easily translated into the framework of temporal logic and model checking. By translating these notions to PCTL, we allow description of vital features that have previously been left out of computational approaches to causal inference, namely the temporal component of the causal relationship as well as explicit description of the sets of conditions comprising a cause. While many of our inferred causes will be spurious, treating the problem of weeding out insignificant causes as a multiple hypothesis testing problem using an empirical null allows us to remain neutral as to the underlying distribution of the data, while still controlling our false discovery rate. This general approach has applications in a variety of areas (e.g. economics, biology, politics), which we have illustrated through two example datasets with very different underlying structures. Supplementary material is available at: http://cs.nyu.edu/~samantha/papers/tlcs.html.

## References

R. Agrawal and R. Srikant. Mining sequential patterns. In *Proc. of the Eleventh International Conference on Data Engineering*, 1995.

C. Baier, E. Clarke, V. Hartonas-Garmhausen, et al. Symbolic model checking for probabilistic processes. In *Proc. 24th International Colloquium on Automata, Languages and Programming*, volume 1256 of *LNCS*. Springer, 1997.

Z. Bar-Joseph. Analyzing time series gene expression data. *Bioinformatics*, 20(16):2493–2503, 2004.

Z. Bozdech, M. Llinás, B. Pulliam, et al. The Transcriptome of the Intraerythrocytic Developmental Cycle of Plasmodium falciparum. *PLoS Biol*, 1(1): e5, 2003.

E. M. Clarke, O. Grumberg, and D. A. Peled. *Model Checking*. MIT Press, 1999.

E. Eells. *Probabilistic Causality*. Cambridge University Press, 1991.

B. Efron. Large-Scale Simultaneous Hypothesis Testing: The Choice of a Null Hypothesis. *Journal of the American Statistical Association*, 99(465):96–105, 2004.

M. Eichler and V. Didelez. Causal reasoning in graphical time series models. In *Proc. of the 23rd Conference on Uncertainty in Artificial Intelligence*, 2007.

N. Friedman, K. Murphy, and S. Russell. Learning the structure of dynamic probabilistic networks. In *Proc. of the 14th Conference on Uncertainty in Artificial Intelligence*, 1998.

N. Friedman, M. Linial, I. Nachman, et al. Using Bayesian Networks to Analyze Expression Data. *Journal of Computational Biology*, 7(3-4):601–620, 2000.

C. W. Granger. Investigating causal relations by econometric models and cross-spectral methods. *Econometrica*, 37(3):424–438, 1969.

H. Hansson and B. Jonsson. A logic for reasoning about time and reliability. *Formal Aspects of Computing*, 6(5):512–535, 1994.

J. Jin and T. T. Cai. Estimating the Null and the Proportion of non-Null effects in Large-scale Multiple Comparisons. *Journal of the American Statistical Association*, 102:495–506, 2006.

C. Langmead, S. Jha, and E. Clarke. Temporal logics as query languages for dynamic bayesian networks: Application to d. melanogaster embryo development. Technical Report CMU-CS-06-159, Carnegie Mellon University, 2006.

C. J. Langmead. Towards inference and learning in dynamic bayesian networks using generalized evidence. Technical Report CMU-CS-08-151, Carnegie Mellon University, 2008.

K. Murphy and S. Mian. Modelling gene expression data using dynamic bayesian networks. Technical report, University of California, Berkeley, CA, 1999.

J. Pearl. *Causality: Models, Reasoning, and Inference*. Cambridge University Press, 2000.

H. Reichenbach. *The Direction of Time*. University of California Press, 1956.

P. Spirtes, C. Glymour, and R. Scheines. *Causation, Prediction, and Search*. MIT Press, 2000.

P. Spirtes, C. Glymour, R. Scheines, et al. Constructing Bayesian Network Models of Gene Expression Networks from Microarray Data. In *Proc. of the Atlantic Symposium on Computational Biology, Genome Information Systems and Technology*, 2001.

P. Suppes. *A probabilistic theory of causality*. North-Holland, 1970.